# Inference-less Density Estimation using Copula Bayesian Networks

**Gal Elidan**
Department of Statistics
Hebrew University
Jerusalem, 91905, Israel
galel@huji.ac.il

## Abstract

We consider learning continuous probabilistic graphical models in the face of missing data. For non-Gaussian models, learning the parameters and structure of such models depends on our ability to perform efficient inference, and can be prohibitive even for relatively modest domains. Recently, we introduced the Copula Bayesian Network (CBN) density model - a flexible framework that captures complex high-dimensional dependency structures while offering direct control over the univariate marginals, leading to improved generalization. In this work we show that the CBN model also offers significant computational advantages when training data is partially observed. Concretely, we leverage on the specialized form of the model to derive a computationally amenable learning objective that is a lower bound on the log-likelihood function. Importantly, our energy-like bound circumvents the need for costly inference of an auxiliary distribution, thus facilitating practical learning of high-dimensional densities. We demonstrate the effectiveness of our approach for learning the structure and parameters of a CBN model for two real-life continuous domains.

## 1 Introduction

Learning multivariate continuous densities is of central importance in a wide range of fields ranging from geology to computational biology. The framework of Bayesian networks (BNs) (Pearl, 1988), geared toward learning such densities in high-dimension, relies on a directed graph structure that encodes independencies that are assumed to hold in the domain. This results in a decomposition of the density into local terms (corresponding to each variable and its parents in the graph) which in turn facilitates relatively efficient inference and learning.

For non-Gaussian continuous graphical models such as those that use a sigmoid Gaussian conditional probability distributions (see, for example, (Koller and Friedman, 2009, Chapter 5)), the maximum likelihood parameters do not have a closed form solution even when the data is fully observed, and estimation of these parameters is carried out using a standard gradient based approach. When some of the observations are missing, learning is computationally daunting for all but the simplest domains since, at every optimization step, computation of the log-likelihood function requires costly inference (high-dimensional integration). To cope with this we typically adopt an expectation maximization approach (Dempster et al., 1977) that relies on an approximate inference method. The variational mean field method (Jordan et al., 1998), for example, makes learning somewhat more practical by decomposing the needed computations according to the network structure. Yet, even with this approximation, parameter learning of non-Gaussian continuous Bayesian networks can be demanding. For many domains learning becomes prohibitive when we also aim to learn the structure of the network, a task that requires a large number of parameter estimates.

Our goal in this work is to overcome these difficulties and perform effective density estimation with partial observations using the Copula Bayesian Network (CBN) density model we recently introduced (Elidan, 2010). Copulas (Nelsen, 2007; Joe, 1997) provide a general framework for the representation of multivariate distributions that separates the choice of marginals and that of the dependency structure. This provides great flexibility and allows us, for example, to construct useful multivariate densities by combining robust and accurate non-parametric univariate estimates (e.g., (Parzen, 1962)) with dependency functions that have a small number of parameters. Yet, despite a dramatic growth in academic and practical interest, copulas are for the most part practical only for relatively low dimensional domains ($< 10$ variables). The CBN model fuses the BN and copula frameworks. Like BNs, it utilizes a directed acyclic graph to encode independencies that are assumed to hold in the domain, leading to a decomposable parameterization. Uniquely, the model relies on local copula func-tions along with an explicit parameterization for the univariate marginals that is shared by the entire network. As demonstrated in Elidan (2010), this can lead to improved generalization of unseen test data in a variety of domains where the data is fully observed.

We present a novel method for density estimation with partial observations that leverages on the specialized form of the CBN model. Starting with the CBN parameterization, we derive an approximate learning objective in the form of a lower bound on the log-likelihood function that is closely related to the variational mean-field lower bound (Jordan et al., 1998). Unlike the standard variational method, by taking advantage of the explicit parameterization of the univariate marginals in the CBN model, we circumvent the need to infer the auxiliary variational distribution in the expectation stage of the EM procedure. As a result, we can perform inference-less estimation, significantly speeding up the learning procedure. In our experimental evaluation in Section 4 this allows us to learn non-Gaussian densities that generalize well in scenarios where BN alternatives proved either ineffective or computationally prohibitive.

The rest of the paper is organized as follows. In Section 2 we provide a brief overview of copulas and copula networks. In Section 3 we present our approach for learning copula networks with partial observations. In Section 4 we apply our approach to two real-life domains. We finish with a discussion and future directions in Section 5.

## 2 Copulas and Copula Bayesian Networks

In this section we briefly review copulas and the recently introduced Copula Bayesian Network model (Elidan, 2010). We start with needed notation.

Let $\mathcal{X} = \{X_1, \ldots, X_N\}$ be a finite set of real random variables and let $F_{\mathcal{X}}(\mathbf{x}) \equiv P(X_1 \leq x_1, \ldots, X_n \leq x_N)$ be a joint (cumulative) distribution function over $\mathcal{X}$, with lower case letters denoting assignment to variables. By slight abuse of notation, we use $F(x_i) \equiv F_{X_i}(x_i) \equiv F(X_i \leq x_i, X_{\mathcal{X}/X_i} = \infty)$. Similarly use $f(x_i) \equiv f_{X_i}(x_i)$ to denote the marginal density function of $X_i$. For a set of variables $\mathbf{Y}$, we use the shorthand $f(\mathbf{y}) \equiv f_{\mathbf{Y}}(\mathbf{y})$.

### 2.1 Copulas

A copula (Sklar, 1959; Nelsen, 2007) is a function that links the marginal distributions together to form a multivariate distribution. Formally,

**Definition 2.1:** Let $U_1, \ldots, U_N$ be real random variables marginally uniformly distributed on $[0, 1]$. A copula function $C : [0, 1]^N \to [0, 1]$ is a joint distribution function

$$C(u_1, \ldots, u_N) = P(U_1 \leq u_1, \ldots, U_N \leq u_N)$$

∎

Interest in copulas has grown dramatically following Sklar's seminal result that captures their expressive power

**Theorem 2.2:** **[Sklar 1959]** *Let $F(x_1, \ldots, x_N)$ be any multivariate distribution over real-valued random variables. Then there exists a copula function such that*

$$F(x_1, \ldots, x_N) = C(F(x_1), \ldots, F(x_N)).$$

*Furthermore, if each $F(x_i)$ is continuous then $C$ is unique.*

The constructive converse which is of central interest from a modeling perspective is also true: since for *any* random variable the cumulative distribution $F(x_i)$ is uniformly distributed on $[0, 1]$, any copula function taking the marginal distributions $\{F(x_i)\}$ as its arguments, defines a valid joint distribution with marginals $F(x_i)$. Thus, copulas are "distribution-generating" functions that allow us to combine *any* univariate representation (e.g. robust non-parametric) with *any* copula function to form a valid joint distribution (see example below).

To derive the joint density $f(\mathbf{x}) = \frac{\partial^N F(\mathbf{x})}{\partial x_1 \ldots \partial x_N}$ from the copula construction, assuming that $F$ has N-order partial derivatives[1] and using the derivative chain rule we have

$$\begin{aligned} f(\mathbf{x}) &= \frac{\partial^N C(F(x_1), \ldots, F(x_N))}{\partial F(x_1) \ldots \partial F(x_N)} \prod_{i=1}^{N} f(x_i) \\ &= c(F(x_1), \ldots, F(x_N)) \prod_i f(x_i), \end{aligned} \quad (1)$$

where $c(F(x_1), \ldots, F(x_N))$, the N-order partial derivative of the copula function $C(F(x_1), \ldots, F(x_N))$, is called the *copula density function*.

**Example 2.3:** A simple copula that we use in this paper and that received significant attention in recent years in the financial community is the elliptical Gaussian copula constructed directly by inverting Sklar's theorem (Embrechts et al., 2003)

$$C(\{F(x_i)\}) = \Phi_\Sigma \left( \Phi^{-1}(F(x_1)), \ldots, \Phi^{-1}(F(x_N)) \right), \quad (2)$$

where $\Phi$ is the standard normal distribution function and $\Phi_\Sigma$ is the zero mean multivariate normal distribution with correlation matrix $\Sigma$. In this paper we consider the simplest variant of this copula where $\Sigma$ has a uniform correlation structure (i.e., when $\Sigma$ has a unit diagonal and a single parameter $\rho$ elsewhere).

To get a sense of the modeling power copulas offer, Figure 1 shows samples from two bivariate joint densities that use the *same* copula function but differ in their marginal form. The first uses normal marginals with a unit mean and variance. The second uses Gamma marginals with the

[1] For any continuous cumulative distribution function $F$ this in fact true almost everywhere

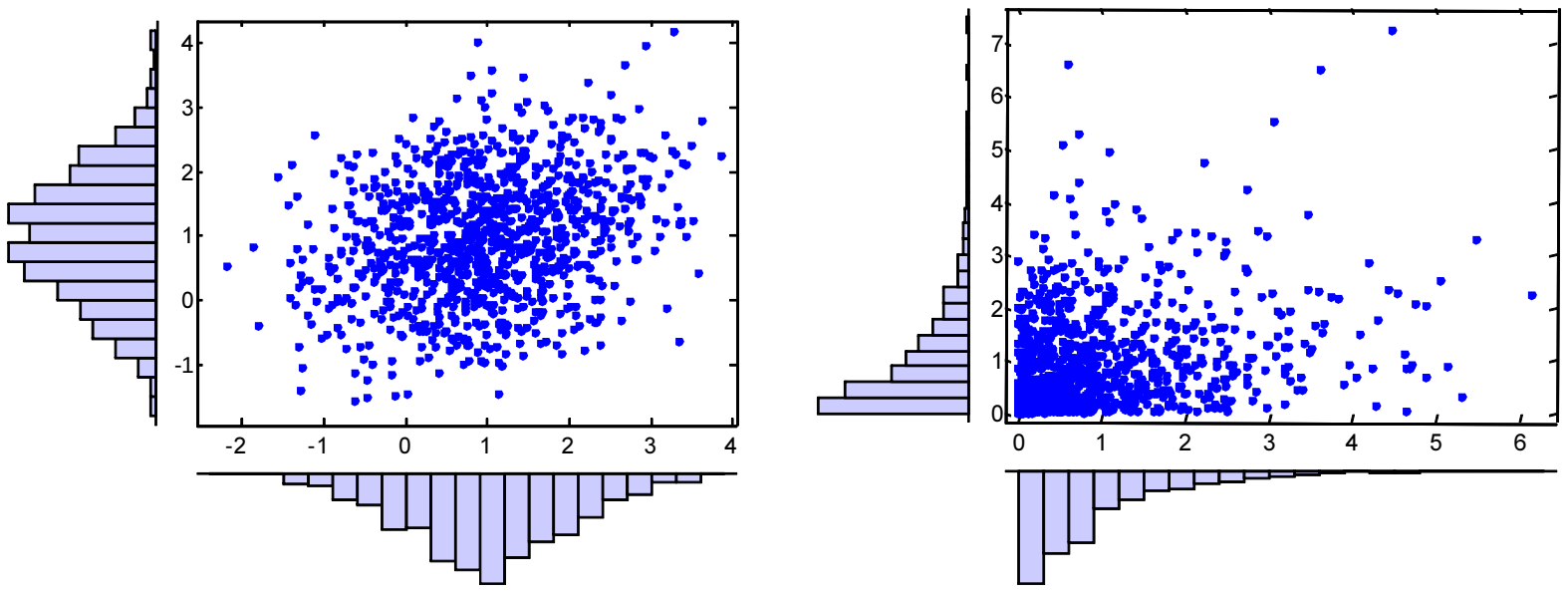

Norma$(1,1)$ marginals  Gamma$(1,1)$ marginals

Figure 1: Demonstration of the modeling flexibility offered by the 2-dimensional Gaussian copula using a correlation matrix with a unit diagonal and a correlation coefficient of 0.25. (left) shows samples from the bivariate density when the marginals are modeled as a normal distribution with a unit mean and variance; (right) shows samples from a bivariate density that uses the *same* copula function and Gamma marginals with the same mean and variance.

same mean and variance, resulting in a substantially different bivariate distribution. More generally and without added computational difficulty, we can also mix and match marginals of different forms.

## 2.2 Copula Bayesian Networks

CBNs, like BNs, rely on a directed acyclic graph (DAG) structure to encode independencies that are assumed to hold in the domain. This graph is joined by a re-parameterization of the conditional density using copula functions, resulting in a compact representation of a a multivariate density where the form of the univariate marginals can be explicitly controlled.

Let $\mathcal{G}$ be a directed acyclic graph whose nodes correspond to the random variables $\mathcal{X} = \{X_1, \ldots, X_N\}$, and let $\mathbf{Pa}_i = \{\mathbf{Pa}_{i1}, \ldots, \mathbf{Pa}_{ik_i}\}$ be the parents of $X_i$ in $\mathcal{G}$. We use $\mathcal{G}$ to encode the independence statements $I(\mathcal{G}) = \{(X_i \perp \mathit{NonDescendants}_i \mid \mathbf{Pa}_i)\}$, where $\perp$ denotes the independence relationship and $\mathit{NonDescendants}_i$ are nodes that are non-descendants of $X_i$ in $\mathcal{G}$.

**Definition 2.4 :** A Copula Bayesian Network (CBN) is a triplet $\mathcal{C} = (\mathcal{G}, \Theta_C, \Theta_f)$ that encodes a joint density $f_{\mathcal{X}}(\mathbf{x})$. $\mathcal{G}$ encodes the independence statements $(X_i \perp \mathit{NonDescendants}_i \mid \mathbf{Pa}_i)$, that are assumed to hold in $f_{\mathcal{X}}(\mathbf{x})$. $\Theta_C$ is a set of local copula density functions $c(F(x_i), F(\mathbf{pa}_{i1}), \ldots, F(\mathbf{pa}_{ik_i}))$ that are associated with the nodes of $\mathcal{G}$ that have at least one parent. In addition, $\Theta_f$ is the set of parameters representing the marginal densities $f(x_i)$ (and distributions $F(x_i)$). The joint density $f_{\mathcal{X}}(\mathbf{x})$ is then parameterized as

$$f_{\mathcal{X}}(\mathbf{x}) = \prod_{i=1}^{N} R_{c_i}\left(F(x_i), F(\mathbf{pa}_{i1}), \ldots, F(\mathbf{pa}_{ik_i})\right) f(x_i).$$

$R_{c_i}$ is a term parameterized by a local copula $c_i$ and that, if $X_i$ has at least one parent in the graph $\mathcal{G}$, is defined to be the following ratio

$$\begin{aligned} & R_{c_i}\left(F(x_i), F(\mathbf{pa}_{i1}), \ldots, F(\mathbf{pa}_{ik_i})\right) \\ \equiv\ & \frac{c_i(F(x_i), F(\mathbf{pa}_{i1}), \ldots, F(\mathbf{pa}_{ik_i}))}{\int c_i(F(x_i), F(\mathbf{pa}_{i1}), \ldots, F(\mathbf{pa}_{ik_i})) f(x_i) dx_i} \\ =\ & \frac{c_i(F(x_i), F(\mathbf{pa}_{i1}), \ldots, F(\mathbf{pa}_{ik_i}))}{\frac{\partial^K C_i(1, F(y_1), \ldots, F(y_K))}{\partial F(y_1) \ldots \partial F(y_K)}}. \end{aligned}$$

When $X_i$ has no parents in the graph $\mathcal{G}$, we define $R_{c_i}\left(F(x_i), F(\mathbf{pa}_{i1}), \ldots, F(\mathbf{pa}_{ik_i})\right)$ to be 1. ∎

It can be shown (Elidan, 2010) that a CBN defines a coherent joint density, and further that the product of local ratio terms $R_{c_i}$ defines a valid joint copula over $\mathcal{X}$. Thus, like other probabilistic graphical models, a CBN takes advantage of the independence assumptions encoded in $\mathcal{G}$ to represent $f_{\mathcal{X}}(\mathbf{x})$ compactly via a product of local terms. Uniquely, a CBN has an explicit marginal form, offering significant modeling advantages in practice (see Elidan (2010) for more details).

Note that in the above definition, the derivative form of the denominator of $R_{c_i}$ is important since for most copula functions it can be computed analytically and integration is not needed. It is also important to note that the univariate distributions are parameterized independently of the local copula functions and are actually shared by several of them.

## 2.3 Learning with Complete Data

Given a complete dataset $\mathcal{D}$ of $M$ instances where all of the variables $\mathcal{X}$ are observed in each instance, the log-likelihood of the data given a CBN model $\mathcal{C}$ is

$$\ell(\mathcal{D} : \mathcal{C}) = \sum_{m=1}^{M} \sum_{i=1}^{N} \left(\log f_i[m] + \log R_{c_i}[m]\right), \quad (3)$$

where $f_i[m]$ is a shorthand for $f(x_i[m])$ and $R_{c_i}[m]$ is similarly a shorthand for the value that the copula ratio $R_{c_i}$ takes in the $m$'th instance.

While the above objective appears to fully decompose according to the structure of the graph $\mathcal{G}$, each marginal dis-tribution $F(x_i)$ actually appears in several copula terms (of $X_i$ and all of its children in $\mathcal{G}$). A solution adopted from the copula community is the Inference Functions for Margins approach (Joe and Xu, 1996), where the marginals are estimated first (see details below). Given $\{F(x_i)\}$, we can then estimate the parameters of each local copula *independently* of the others, using a standard gradient based approach.

To learn the structure $\mathcal{G}$ of a CBN, we use the score-based approach commonly used in the BN community. Briefly, starting with the empty network, we perform a greedy search for a beneficial structure via evaluation of local modifications to the current structure (add/delete/reverse an edge). The search is guided by a model selection score that balances the log-likelihood of the model and its complexity

$$\text{score}(\mathcal{G} \;:\; \mathcal{D}) = \ell(\mathcal{D} : \hat{\theta}, \mathcal{G}) - \text{Pen}_M(\mathcal{G}),$$

where $\hat{\theta}$ are the maximum likelihood parameters that correspond to the graph $\mathcal{G}$, and $\text{Pen}_M(\mathcal{G})$ is a penalty function that depends on the structure of the graph and number of instances $M$ in $\mathcal{D}$ but not on the data itself. A common model selection score is the Bayesian Information Criterion (Schwarz, 1978) for which $\text{Pen}_M(\mathcal{G}) = \frac{1}{2}\log(M)|\Theta_{\mathcal{G}}|$, where $|\Theta_{\mathcal{G}}|$ is the number of free parameters associated with the graph structure $\mathcal{G}$. See Koller and Friedman (2009) for more details on this standard model selection approach in the context of regular BNs.

# 3 Inference-less Learning

We now consider the estimation of a CBN model when some of the values in the dataset $\mathcal{D}$ of $M$ instances are missing. As we will show, in this scenario CBNs do not only offer a generalization advantage but also give rise to computational advantages in the form of relatively efficient inference-less learning.

## 3.1 Estimation of the Univariate Marginals

A central motivation for the use of CBNs and copulas in general is that estimation of the univariate marginals is relatively straightforward and can be carried out robustly. To examine whether this assumptions holds in practice, we evaluate the effect of missing (at random) data on the marginals for the two very different real-life domains we consider in our experimental evaluation.

To estimate $F(x_i)$ and the univariate densities $f(x_i)$ we use a standard kernel-based approach (Parzen, 1962). Given $x_i[1], \ldots, x_i[M]$ i.i.d. samples of a random variable $X_i$, the kernel density estimate of its probability density function is

$$\hat{f}_h(x) = \frac{1}{Mh}\sum_{i=1}^{M} K\left(\frac{x - x_i}{h}\right)$$

where $K$ is some kernel function and $h$ is a smoothing parameter called the bandwidth. Qualitatively, the method approximates the distribution by placing small "bumps" (determined by the kernel) at each data point. Thus, higher density values will result in regions where there is a concentration of data samples. In this work we use the standard and mathematically convenient Gaussian kernel (see, for example, Bowman and Azzalini (1997) for details).

Figure 2 shows the dependence of the non-parametric univariate marginal estimates on the amount of data observed. Appealingly, with as little as $5\%$ of the data, the marginal estimation is reasonable. It is in fact quite accurate for the slightly better but still aggressively pessimistic scenario where only $25\%$ of the variables are observed in each instance. These results suggest that in the face of missing at random data, we should simply use the observed values for estimation of the univariate marginal. Thus, just as in the complete data scenario, we turn to the estimation of the local copula functions of the CBN model assuming given (or estimated) univariate marginals.

## 3.2 Estimation of the Local Copula Functions

When some of the values are missing in $\mathcal{D}$, given the univariate marginals, the log-likelihood function becomes

$$\ell(\mathcal{D} : \mathcal{C}) = \sum_{m=1}^{M} \log \int \left[\prod_{i=1}^{N} R_{c_i}[m]\hat{f}_i[m]\right] d\mathbf{H}_m, \qquad (4)$$

where $\mathbf{H}_m$ are the variables not observed in the $m$'th instance, and $\hat{f}$ are the given or estimated univariate marginals. $\hat{f}_i[m]$ is a shorthand for the estimated $\hat{f}(x_i)$ where $x_i$ takes the value of $X_i$ in the $m$'th instance if observed or the value assigned to $X_i$ in the integral if not. Similarly, $R_{c_i}[m]$ is a shorthand for the value that $R_{c_i}\left(F(x_i), F(\mathbf{pa}_{i1}), \ldots, F(\mathbf{pa}_{ik_i})\right)$ takes given the assignment to $\mathbf{H}_m$ in the integral and the observed variables in the $m$'th instance.

As is the case for the standard non-Gaussian BN log-likelihood function, evaluation and maximization of Eq. (4) is difficult: since each variable appears in several local copula quotient terms, the log-likelihood does not decompose according to the graph's structure as is the case when the data is fully observed. A common solution is to adopt an expectation maximization (EM) approach (Dempster et al., 1977). This, however, requires that we compute, for each instance $m$, the posterior $f(\mathbf{H}_m \mid O[m], \mathcal{M})$ of the hidden variables in the instance $\mathbf{H}_m$ given the observed ones $O[m]$ and the current model $\mathcal{M}$. For all but Gaussian continuous graphical models, this task can be quite costly as it requires high dimensional integration.[2]

An appealing alternative is to construct a workable lower bound on the log-likelihood function and optimize the

[2] We also tried an EM approach using a Gibbs-like Metropolis-Hastings inference procedure (see (Koller and Friedman, 2009, Chapter 14)). However, the approach proved computationally prohibitive even for our smaller 12 variables dataset.

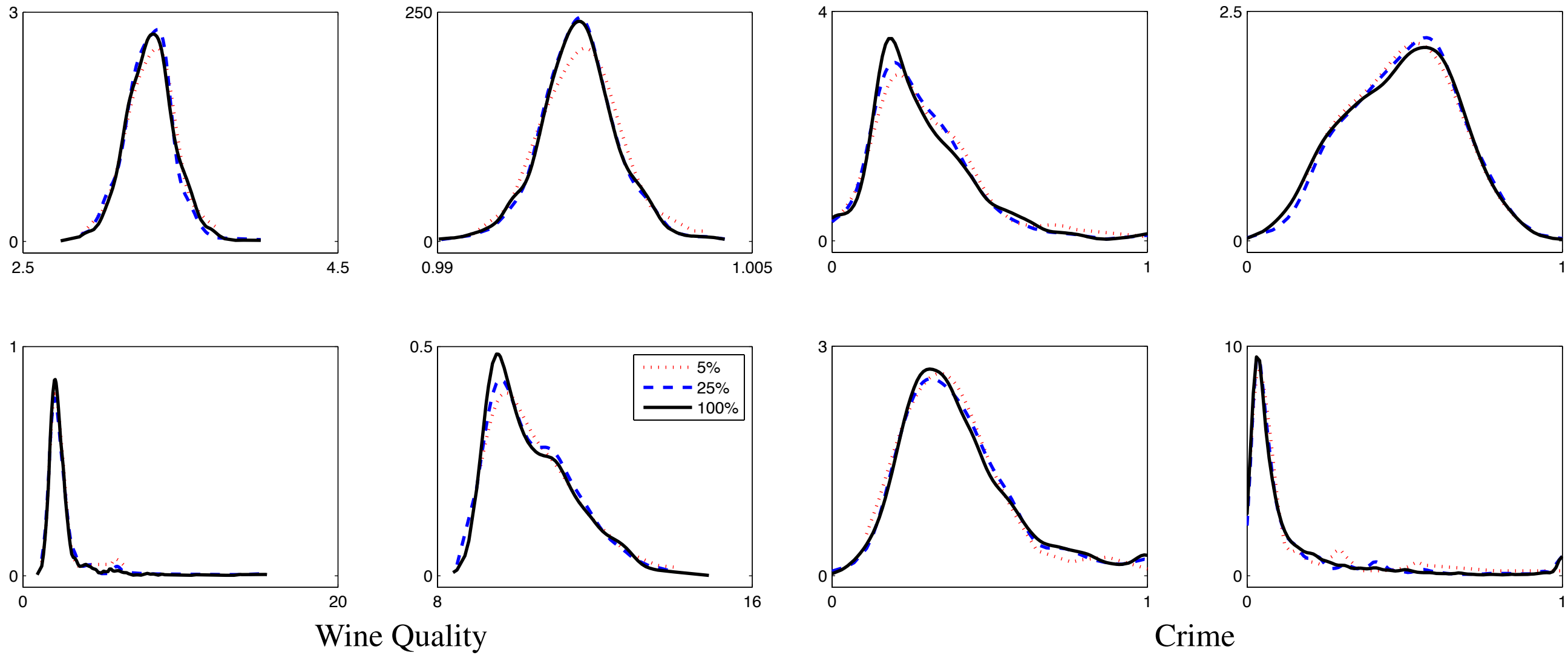

Figure 2: Marginal density functions for 4 randomly chosen variables from the **Wine** and **Crime** datasets (see Section 4) with 5% of the values observed (dotted red), 25% of the values observed (dashed blue) and all values observed (solid black). The robustness of the marginal density estimates is evident even when 95% of the values are missing.

parameters with respect to this bound. Appealingly, a straightforward way to do so without the tools of variational inference arises from the fact that CBNs contain an explicit representation of the univariate marginals:

**Theorem 3.1:** *Let $\mathcal{C}$ be a copula network model over $\mathcal{X}$ and $\mathcal{D}$ be a dataset of $M$ instances. Then the log-likelihood function can be bounded from below using*

$$\ell(\mathcal{D}:\mathcal{C}) \geq \sum_{m=1}^{M}\sum_{i=1}^{N}\int\left[\left(\prod_{H\in\mathbf{H}_m^i}\hat{f}(h)\right)\log\left(R_{c_i}[m]\right)\right]d\mathbf{H}_m^i + \sum_{m=1}^{M}\sum_{X_i\notin\mathbf{H}_m}\log\hat{f}(x_i[m]), \quad (5)$$

*where we use $\mathbf{H}_m^i = \mathbf{H}_m \cap \{X_i, \mathbf{Pa}_i\}$ to denote the variables not observed in the $i$'th family in the $m$'th instance.*

**Proof:** We start by rewriting the term of the log-likelihood function of Eq. (4) that corresponds to the $m$'th instance:

$$\ell_m(\mathcal{D}:\mathcal{C}) = \log\int\left[\left(\prod_{i=1}^{N}R_{c_i}[m]\right)\left(\prod_{H\in\mathbf{H}_m}\hat{f}(h)\right)\right]d\mathbf{H}_m + A_m$$

where the constant $A_m$ absorbs the marginals of the observed variables and equals $\sum_{X_i\notin\mathbf{H}_m}\log\hat{f}(x_i[m])$ (last term in Eq. (5)). Now, since $\hat{f}(h)$ is a density, we can apply Jensen's inequality repeatedly for each $H\in\mathbf{H_m}$ and write

$$\begin{aligned}\ell_m(\mathcal{D}:\mathcal{C}) &\\ \geq & \int\left[\left(\prod_{H\in\mathbf{H}_m}\hat{f}(h)\right)\log\left(\prod_{i=1}^{N}R_{c_i}[m]\right)\right]d\mathbf{H}_m + A_m \\ = & \sum_{i=1}^{N}\int\left[\left(\prod_{H\in\mathbf{H}_m}\hat{f}(h)\right)\log\left(R_{c_i}[m]\right)\right]d\mathbf{H}_m + A_m.\end{aligned}$$

Finally, since $R_{c_i}[m]$ does not depend on variables in $\mathbf{H}_m$ that are not in $\mathbf{H}_m^i$, we can write

$$\begin{aligned}&= \sum_{i=1}^{N}\left(\prod_{H\notin\mathbf{H}_m^i}\int\hat{f}(h)dH\right)\times \\ &\quad\left(\int\left[\left(\prod_{H\in\mathbf{H}_m^i}\hat{f}(h)\right)\log\left(R_{c_i}[m]\right)\right]d\mathbf{H}_m^i\right) + A_m \\ &= \sum_{i=1}^{N}\int\left[\left(\prod_{H\in\mathbf{H}_m^i}\hat{f}(h)\right)\log\left(R_{c_i}[m]\right)\right]d\mathbf{H}_m^i + A_m,\end{aligned}$$

∎

The last line follows from the fact that $\hat{f}(h)$ is a density.

Our lower bound, similarly to energy-based variational approximations (see next section) decomposes according to the network structure and estimation can now be carried out independently for each local copula. Importantly, as we discuss in the next section, computation of our bound is significantly more efficient than its variational counter-part, facilitating practical high-dimensional learning.

### 3.3 A Variational Energy Perspective

Given a set of observed variables $\mathbf{Y} \subset \mathcal{X}$, we are interested in computing $\tilde{f}_{\mathcal{X}}(\mathbf{x}) = f(\mathcal{X} = \mathbf{x} \mid \mathbf{Y} = \mathbf{y})$. In this case, even for BNs where computation of $f_{\mathcal{X}}(\mathbf{x})$ (with $\mathbf{x}$ a complete instance) is straightforward, computation of $\tilde{f}_{\mathcal{X}}(\mathbf{x}) = f_{\mathcal{X}}(\mathbf{x})/f(\mathbf{Y} = \mathbf{y})$ is typically difficult, and amounts to the computation of the likelihood function $f(\mathbf{Y} = \mathbf{y})$. In the variational approach for approximate inference (Jordan et al., 1998), we use an auxiliary density $q_{\mathcal{X}}(\mathbf{x})$ of a convenient form to approximate the less tractable $\tilde{f}_{\mathcal{X}}(\mathbf{x})$. For any choice of $q_{\mathcal{X}}(\mathbf{x})$ we attempt to find $q^*_{\mathcal{X}}(\mathbf{x})$ that minimizes the Kullback-Leibler divergence (Kullback and Leibler, 1951) between $q_{\mathcal{X}}(\mathbf{x})$ and our target $\tilde{f}_{\mathcal{X}}(\mathbf{x})$. Excluding terms that do not depend on $q_{\mathcal{X}}(\mathbf{x})$, we can equivalently maximize the following functional

$$\mathcal{E}[q_{\mathcal{X}}(\mathbf{x}), f_{\mathcal{X}}(\mathbf{x})] = \boldsymbol{E}_{q_{\mathcal{X}}(\mathbf{x})}[\log f_{\mathcal{X}}(\mathbf{x}) - \log q_{\mathcal{X}}(\mathbf{x})]. \quad (6)$$

Eq. (6), called the energy functional, is also (for any $q_{\mathcal{X}}(\mathbf{x})$) a lower bound on the log-likelihood function and is tight only if $q_{\mathcal{X}}(\mathbf{x})$ equals to $\tilde{f}_{\mathcal{X}}(\mathbf{x})$. In the mean-field approximation $q_{\mathcal{X}}(\mathbf{x})$ is chosen to have a simple product of marginals form $q_{\mathcal{X}}(\mathbf{x}) = \prod_i q(x_i)$. We can then search for $q^*_{\mathcal{X}}(\mathbf{x})$ that (locally) maximizes the energy functional via a straightforward iterative procedure (see Jordan et al. (1998); Koller and Friedman (2009), for more details).

Interestingly, although our lower bound was derived from a direct application of Jensen's inequality to the CBN density without *any* additional assumptions, it turns out that our objective is closely related to the mean field free energy one. Using $\mathbf{o}[m]$ to denote the observed values in the $m$'th instances, if one chooses

$$q^*(\mathcal{X} = (\mathbf{o}[m], \mathbf{h}[m])) = \prod_{H \in \mathbf{H_m}} \hat{f}(h[m]) \prod_{O \notin \mathbf{H_m}} \hat{f}(o[m]) \quad (7)$$

then, since for the copula network model $\log f(\mathbf{x}) - \log q^*(\mathbf{x}) = \sum_i \log R_{c_i}\left(F(x_i), F(\mathbf{pa}_{i1}), \ldots, F(\mathbf{pa}_{ik_i})\right)$, the energy functional for the $m'th$ instance is $\mathcal{E}[q^*_{\mathcal{X}}(\mathbf{x}), f_{\mathcal{X}}(\mathbf{x})] = \sum_i \boldsymbol{E}_{q^*_{\mathcal{X}}(\mathbf{x})}[\log R_{c_i}[m]]$. In this case, the mean field energy functional is equal to our lower bound of Eq. (5) minus $\sum_{X_i \notin \mathbf{H}_m} \log \hat{f}(x_i)$ for each instance $m$.

What are the implications of this variational perspective of our lower bound? We start by noting that the additional term in our bound does not depend on the local copula functions. Thus, given $q^*_{\mathcal{X}}(\mathbf{x})$, for optimization purposes we can simply use the standard energy functional. As an approximation to the log-likelihood function, our objective can be either inferior or superior to the energy functional, depending on the peakedness of the marginal distributions. In our experimental evaluation we did not observe any conclusive advantage to either objective.

The obvious question is then why choose $q^*_{\mathcal{X}}(\mathbf{x})$ as we did and not the one that maximizes $\mathcal{E}[q_{\mathcal{X}}(\mathbf{x}), f_{\mathcal{X}}(\mathbf{x})]$ and is thus guaranteed to provide a better lower bound for the log-likelihood function.[3] The answer to this question is two-fold. First, finding the optimum of $\mathcal{E}[q_{\mathcal{X}}(\mathbf{x}), f_{\mathcal{X}}(\mathbf{x})]$ is difficult and the mean-field iterations only provide a *locally* optimal solution. Second and most importantly, our choice of $q^*_{\mathcal{X}}(\mathbf{x})$ as the product of the given (or estimated) marginals circumvents the need for the mean field iterations as the needed marginals are an explicit part of the CBN representation. For continuous non-Gaussian models where mean-field inference can be quite costly, our approach can be crucial for facilitating learning of high-dimensional CBNs (see Section 4).

## 4 Experimental Evaluation

We assess the effectiveness of our approach for density estimation by comparing CBNs and BNs learned for two markedly different real-life datasets:

- **Wine Quality**. 11 physiochemical properties and a sensory quality (0-10) variable for the red variant of the Portuguese "Vinho Verde" wine (Cortez et al., 2009). Dataset includes measurements from 1599 tastings over a period of three years.
- **Crime**. The communities and crime dataset from the UCI machine learning repository including 100 variables ranging from mean household size to percentage of kids born outside of a marriage. Included are values for 1994 communities across the U.S.

For BNs, we use a linear Gaussian conditional density where each variable is distributed as a Gaussian whose mean is a linear function of its parent's values. For this model, exact inference can be carried out in closed form. We also tried learning Bayesian Networks with a more expressive Sigmoid Gaussian conditional density for each variable. This, however, led to inferior generalization performance for the 12 variable **Wine** dataset (results not reported for clarity) and was computationally prohibitive for the 100 variable **Crime** dataset. For CBNs, we use the uniform correlation Gaussian copula of Eq. (2). We use standard Gaussian kernel density estimation for the univariate densities as described in Section 3.1. The structure of both the Bayesian Network and CBN models was learned using the same greedy procedure described in Section 2.3

We start with a quantitative evaluation of the average log-probability per instance of the train and test data given the learned model BN and CBN models. All results reported below are for 10 random splits into equal train/test sets. Variables were randomly and independently hidden in each instance. All figures show the average performance as well as the $10 - 90\%$ range (error bars).

[3] Note that such a $q_{\mathcal{X}}(\mathbf{x})$ generally *does not* equal to the product of marginals of $f_{\mathcal{X}}(\mathbf{x})$ and is in fact more peaked. See, for example, Koller and Friedman (2009), Chapter 11.

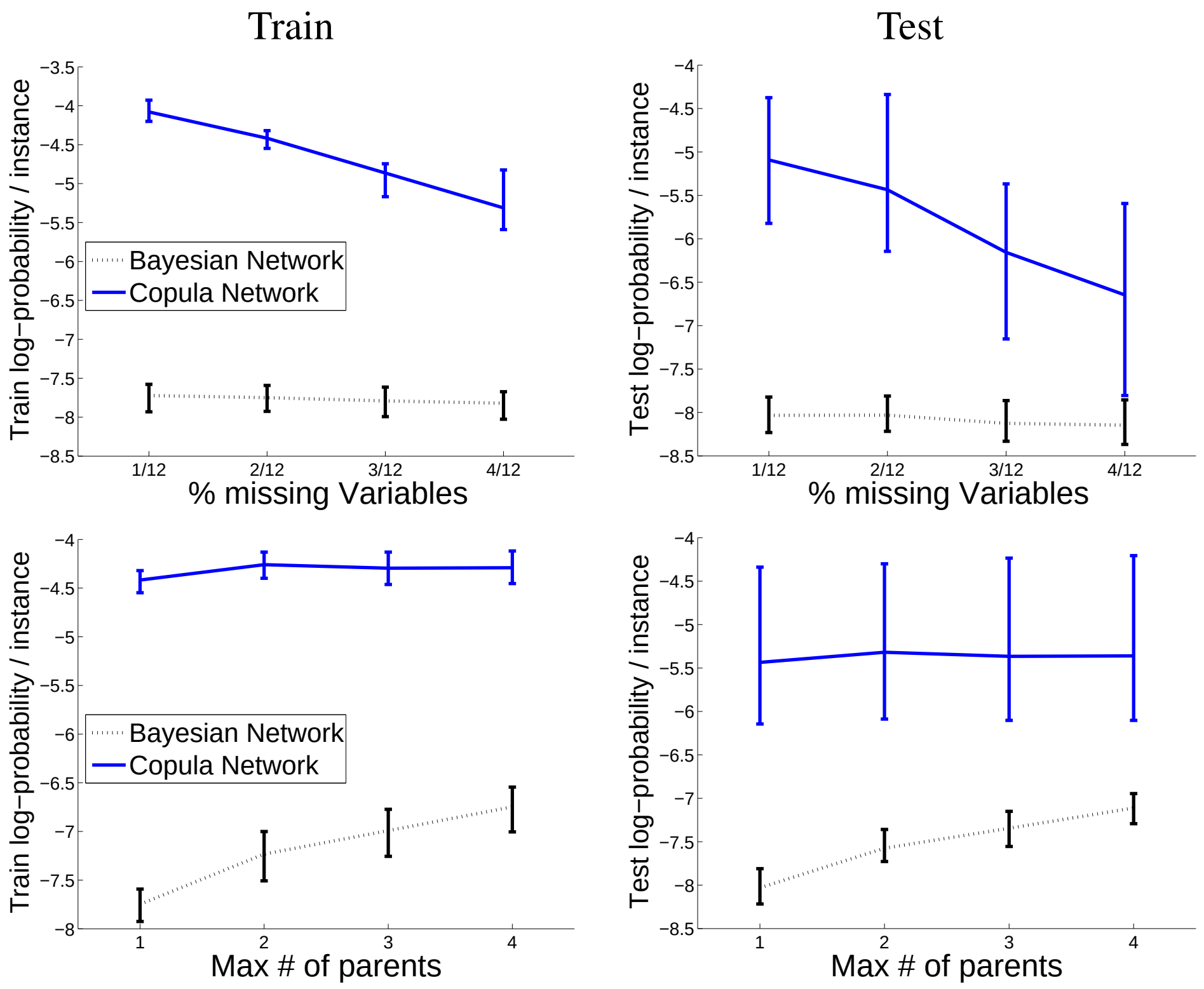

Figure 3: Train (left) and test (right) set performance for the 12 variable **Wine** dataset of a linear Gaussian BN and a CBN with a single parameter uniform correlation Gaussian local copula (the performance of the Sigmoid BN model was slightly inferior to the linear BN while taking an order magnitude longer to learn). Shown is the average log-probability per instance of 10 random runs as well as the $10 - 90\%$ range (error-bars). (top) shows performance as a function of the fraction of missing variables for tree structured networks; (bottom) shows performance as a function of the number of parents allowed for each variable in the network structure with two values randomly hidden in each instance. The structure for all models was learned using a standard greedy procedure and the BIC model selection score.

Figure 3(top) shows the train and test performance (y-axis) for the **Wine** dataset as a function of the fraction of missing values in each instance, where both the CBN and BN models were constrained to a tree structure. The superiority of the copula network model is evident and consistent. Note that the advantage is quite significant as an improvement of 1 in the y-axis is equal to *each* test instance being twice as likely on average. Not surprisingly, as the number of missing values in each instance grows, the CBN models suffers to a greater extent since, unlike the BN model, it relies on an approximation of the log-likelihood learning objective. Yet, even when $25\%$ of the values are not observed, the advantage of the CBN model is substantial.

To show that the advantage of the CBN model does not depend on the fact that the model was constrained to a tree structure, Figure 3(bottom) compares the train and test performance as a function of the number of parents allowed in the network, with $2/12$ values randomly hidden in each instance. Once again, the advantage of the CBN model is evident and consistent. Note that the BN benefits to a greater extent from the increase in the model complexity than the CBN model. This should not come as a surprise since the CBN model has a single parameter for each variable regardless of the number of parents allowed, while the number of parameters for each variable in the linear Gaussian BN equals to the number of parents of the variable in the network plus 2. Obviously, more expressive copulas may lead to greater generalization advantages.

To get a qualitative sense of the advantage of the CBN model, Figure 4 compares empirical values from the test data with samples generated for pairs of variables from the learned BN and CBN models. For the 'physical density' and 'alcohol' variables (top), the samples generated from the CBN model (middle) are somewhat better than those generated from the BN model (right) when compared to the empirical distribution (left), but not dramatically so. However, for the 'residual sugar' and 'physical density' pair (bottom), where the empirical dependence is far from Gaussian, the advantage of the CBN representation clearly manifests. We recall that the CBN model uses a simple Gaussian copula so that the advantage is rooted in the distortion of the input to the copula created by the kernel-based univariate representation. With a choice of a domain specific and more expressive copula, we can expect further qualitative and quantitative advantages.

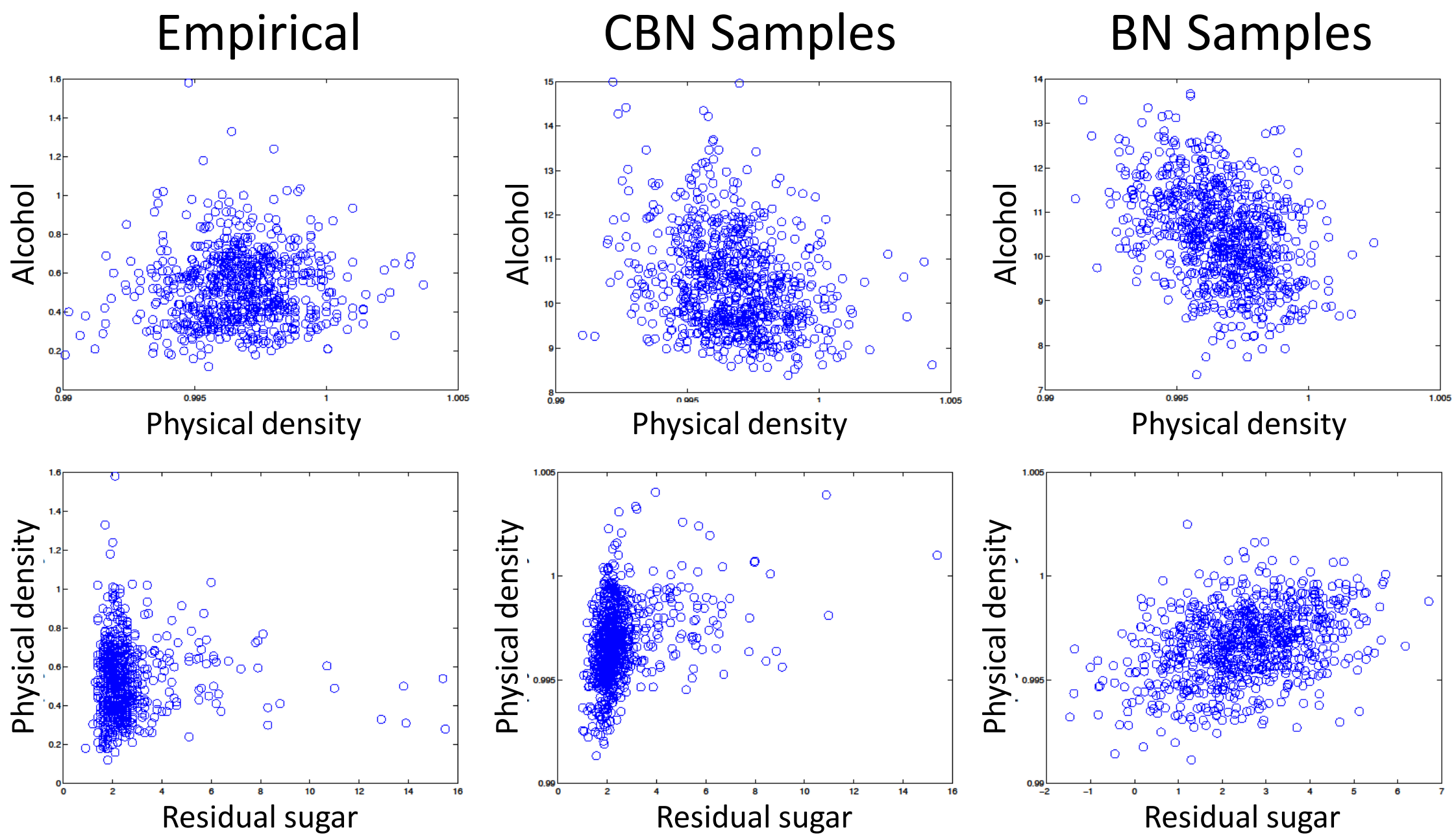

Figure 4: Qualitative illustration of the dependency learned for the **Wine** dataset for two pairs of variables: 'physical density' and 'alcohol level' (top); 'residual sugar' and 'physical density' (bottom). Compared is the empirical distribution of values in the test data (left) with samples generated from the learned CBN (middle) and BN (right) models. To eliminate the effect of differences in structure, the CBN model was forced to use the structure learned for the BN model which contains the network fragment 'residual sugar' $\rightarrow$ 'physical density' $\rightarrow$ 'alcohol level'.

Figure 5 shows the train and test performance (y-axis) for the significantly more challenging 100 variable **Crime** dataset as a function of the fraction of missing variables in each instance. With 5% and 10% missing values in each instances, the CBN model still performs significantly better both on train and test data. However, as the number of missing values increases, performance degrades and with 20% missing values it is actually better to use a simple linear Gaussian BN. This should not come as a surprise as the BN model has an important a-priori advantage: exact inference can be carried out in closed form making estimation more robust in the face of missing data. The lesson here is general - naive models with exact inference can overcome "better" ones when many values are missing and the quality of inference becomes crucial. That said, it is likely that the CBN model can do better if a copula that is better fitted to the application is used (recall that in here, for demonstrative purposes, we used one of the simplest copula functions in the literature). Indeed, part of the strength of the CBN model is that unlike BNs where the choice of the parameterization has a significant impact on our ability to use the model, the vast majority of copula functions are equally "friendly" in terms of computations.

## 5 Discussion and Future Work

We presented a novel method for multivariate continuous density estimation with missing at random data. By leveraging on the unique form of Copula Bayesian Networks, we derived an energy-like lower-bound on the log-likelihood function that can then be used as an approximate learning objective. Importantly, our bound does not require costly inference and facilitates estimation of complex continuous densities in practice. We applied our approach to two real-life scenarios where a non-Gaussian BN model was either ineffective or computationally prohibitive.

Recently, other works have considered a combination of copulas and graphical models. Kirshner (2007) suggested a mixture of tree construction of bivariate copulas. Liu et al. (2010) describe as undirected non-parametric framework that is specific to the multivariate Gaussian copula. Aside from the differences in representation from CBNs, both of these works, as indeed the vast majority of copula literature, do not consider the general (non-Gaussian) problem of estimation in the face of missing data, which in turn relies on our ability to perform posterior computations.

To the best of our knowledge, ours is the first general purpose approach for copula-based density estimation in the face of missing (at random) data. Importantly, the decomposable form of our learning objective *does not* depend on the specific copula used. In addition, by taking advantage of the CBN representation, our procedure is *more* efficient than a similar one for standard non-linear BNs. Thus, the CBN model is unique in that greater expressiveness and improved generalization do not necessarily come at the cost of computational demands. We believe that application oriented exploration of different local copula functions will further enhance the ramifications of this phenomenon.

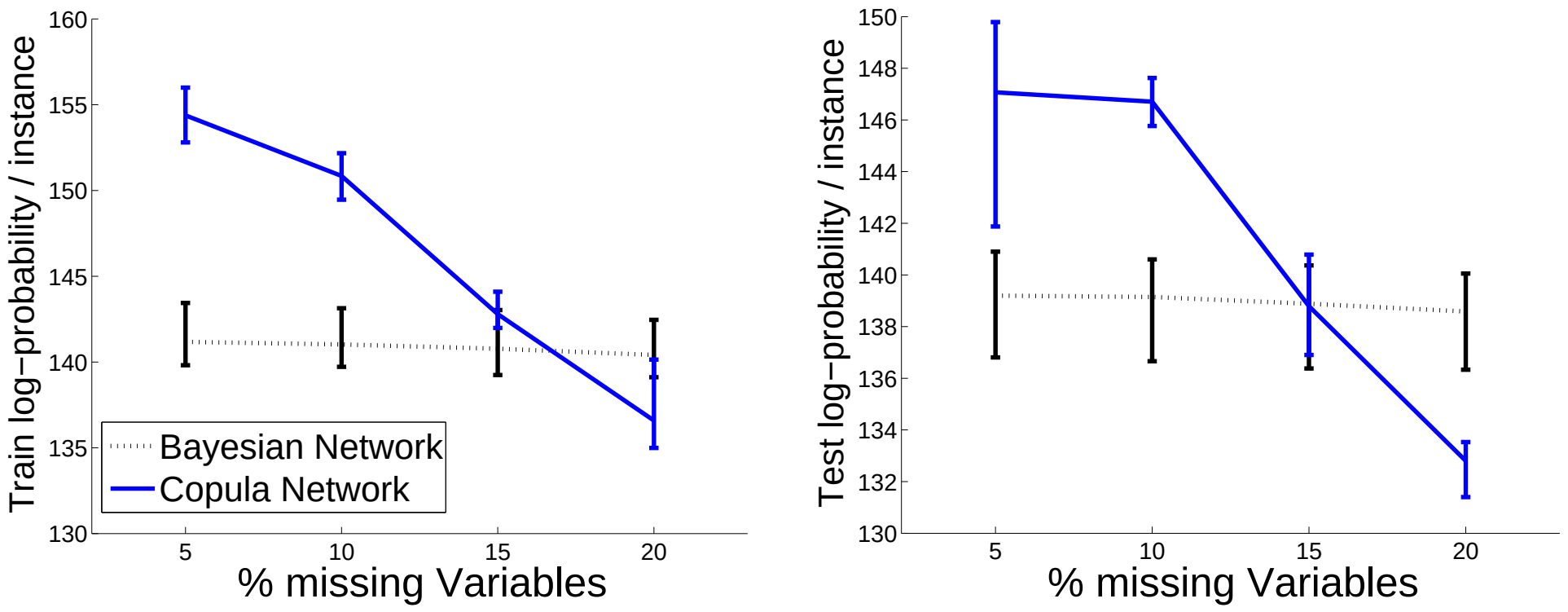

Figure 5: Train (left) and test (right) set performance for the 100 variable **Crime** dataset of a linear Gaussian BN and a CBN with a single parameter uniform correlation Gaussian local copula. Learning a Sigmoid BN model for this dataset proved computationally prohibitive. Shown is the average log-probability per instance of 10 random runs as well as the $10-90\%$ range (error-bars) as a function of the fraction of missing variables.

An obvious question of interest is whether the CBN model provides computational advantages if we require more accurate posterior approximations. A related question is how to adapt our approach to the scenario where the pattern of missing values is not random, and in particular in the face of completely hidden variables. Another direction of interest is the possibility of leveraging on the copula parameterization in orthogonal directions. For example, it would be useful to see whether structure learning can be made substantially more efficient by applying approximate-before-exact model selection approaches (Friedman et al., 1999; Elidan et al., 2007) to the Copula Bayesian Network model.

## Acknowledgments

We thank N. Friedman and A. Globerson for their useful comment on earlier draft of this manuscript. G. Elidan was supported by an Alon fellowship.